# Examining Users' Behavioural Intention to Use OpenClaw Through the Cognition–Affect–Conation Framework

Yiran Du
University of Cambridge
yd392@cam.ac.uk

**Abstract**
This study examines users' behavioural intention to use OpenClaw through the Cognition–Affect–Conation (CAC) framework. The research investigates how cognitive perceptions of the system influence affective responses and subsequently shape behavioural intention. Enabling factors include perceived personalisation, perceived intelligence, and relative advantage, while inhibiting factors include privacy concern, algorithmic opacity, and perceived risk. Survey data from 436 OpenClaw users were analysed using structural equation modelling. The results show that positive perceptions strengthen users' attitudes toward OpenClaw, which increase behavioural intention, whereas negative perceptions increase distrust and reduce intention to use the system. The study provides insights into the psychological mechanisms influencing the adoption of autonomous AI agents.

## 1. Introduction

Artificial intelligence (AI) technologies have increasingly become integrated into digital systems, reshaping how individuals interact with information, services, and automated tools. Recent advances in generative AI have expanded the capabilities of intelligent systems beyond information retrieval to include task automation and decision support (Granić, 2024; Jung & Jo, 2025). In particular, the emergence of autonomous AI agents marks a new stage in AI development, allowing systems to independently plan and execute multi-step tasks on behalf of users. Despite these technological advancements, users' adoption of AI systems remains strongly influenced by their perceptions, attitudes, and concerns regarding system capabilities, reliability, and potential risks (Afroogh et al., 2024; Schwesig et al., 2023).

OpenClaw is an emerging open-source AI agent framework that enables autonomous task execution by integrating large language models with external tools and system interfaces (Shan et al., 2026; Manik & Wang, 2026). Unlike conventional conversational AI systems that primarily generate text responses, OpenClaw can interact with files, application programming interfaces, and operating-system commands to perform complex workflows (Dong et al., 2026). This design reflects the growing shift toward action-oriented AI assistants capable of automating practical tasks and augmenting human productivity. However, the increasing autonomy of such systems also introduces concerns related to privacy, transparency, and potential risks, which may influence users' perceptions and willingness to adopt AI agents (Borjigin et al., 2026; Herriger et al., 2025).

Although previous studies have examined user acceptance of conversational AI and generative AI technologies, research focusing specifically on autonomous AI agent frameworks such as OpenClaw remains scarce. As one of the emerging platforms enabling agent-based automation, OpenClaw has attracted growing attention in recent technical discussions, yet empirical studies investigating users' acceptance of this system are still limited. To address this gap, this study applies the Cognition–Affect–Conation (CAC) framework, which explains how cognitive evaluations shape affective responses and subsequently influence behavioural intention (Zeng et al., 2023; Qaisar et al., 2024). By examining both enabling and inhibiting perceptions, the study seeks to provide early empirical insights into the factors influencing users' behavioural intention to use OpenClaw.

## 2. Literature Review

### 2.1 OpenClaw

OpenClaw (formerly Clawdbot and Moltbot*)* is an emerging open-source AI agent framework designed to execute real tasks on behalf of users rather than merely generating text responses (Shan et al., 2026). Unlike conventional conversational systems, OpenClaw integrates large language models with external tools and system interfaces, enabling it to interact with files, APIs, messaging platforms, and operating-system commands (Manik & Wang, 2026). This design allows the agent to autonomously plan and complete multi-step workflows, positioning OpenClaw as part of the growing paradigm of action-oriented AI assistants that extend beyond passive information retrieval (Dong et al., 2026).

The architecture and functionality of OpenClaw illustrate the increasing convergence between conversational AI and digital automation (T. Chen et al., 2026). By enabling AI agents to perform practical operations such as managing data, retrieving online information, or automating routine workflows, OpenClaw demonstrates the potential of agentic AI systems to augment human productivity and decision-making (E. Chen et al., 2026). However, the deployment of such systems also introduces challenges, including usability constraints, trust considerations, perceived risks related to system autonomy, and concerns regarding privacy and security (Borjigin et al., 2026). These factors may significantly influence how users perceive and adopt agent-based AI technologies.

Given the novelty and functional complexity of OpenClaw, understanding how potential users evaluate and accept such systems is critical. While prior research has extensively examined adoption of conversational AI and intelligent assistants, empirical evidence regarding user acceptance of autonomous AI agents remains limited (Basu, 2026). Therefore, examining users' behavioural intention to use OpenClaw is necessary to identify the determinants that influence adoption, inform system design, and provide insights into how users respond to emerging agentic AI technologies in real-world contexts.

### 2.2 Cognition–Affect–Conation Framework

The Cognition–Affect–Conation (CAC) framework provides a theoretical lens for explaining how individuals develop behavioural intentions through a sequential psychological process (Zeng et al., 2023). Originating from attitude theory in social psychology, the framework proposes that individuals' responses toward an object or technology evolve through three stages: cognitive evaluation, affective reaction, and conative intention (Qaisar et al., 2024). Cognition represents individuals' beliefs and perceptions regarding an object, affect reflects the emotional responses formed from those beliefs, and conation refers to the behavioural intention or action that results from these evaluations (Zhou & Zhang, 2024).

In the context of technology adoption, the CAC framework suggests that users initially form cognitive evaluations of a system based on perceived attributes or characteristics of the technology (Zhou & Zhang, 2025). These cognitive assessments subsequently shape affective responses, such as favourable attitudes or negative emotional reactions. The affective responses then influence the conative stage, where individuals develop intentions regarding whether they will adopt or use the technology. This sequential mechanism has been widely applied in information systems research to explain how users' perceptions and emotions jointly influence behavioural intention (Zhou & Wang, 2025).

Following the CAC logic, this study conceptualises users' perceptions of OpenClaw as the cognitive stage. Specifically, enabling perceptions include perceived personalisation, perceived intelligence, and relative advantage, while inhibiting perceptions include privacy concern, algorithmic opacity, and perceived risk. These cognitive evaluations shape users' affective responses toward the technology, represented by attitude and distrust. Attitude captures users' overall favourable evaluation of the technology, whereas distrust reflects negative affective reactions arising from perceived risks and uncertainties. In turn, these affective responses influence users' behavioural intention to use OpenClaw (see Figure 1). By applying the CAC framework, this study explains how both positive and negative perceptions of autonomous AI agents translate into users' adoption intentions.

**Figure 1. The Conceptual Model**

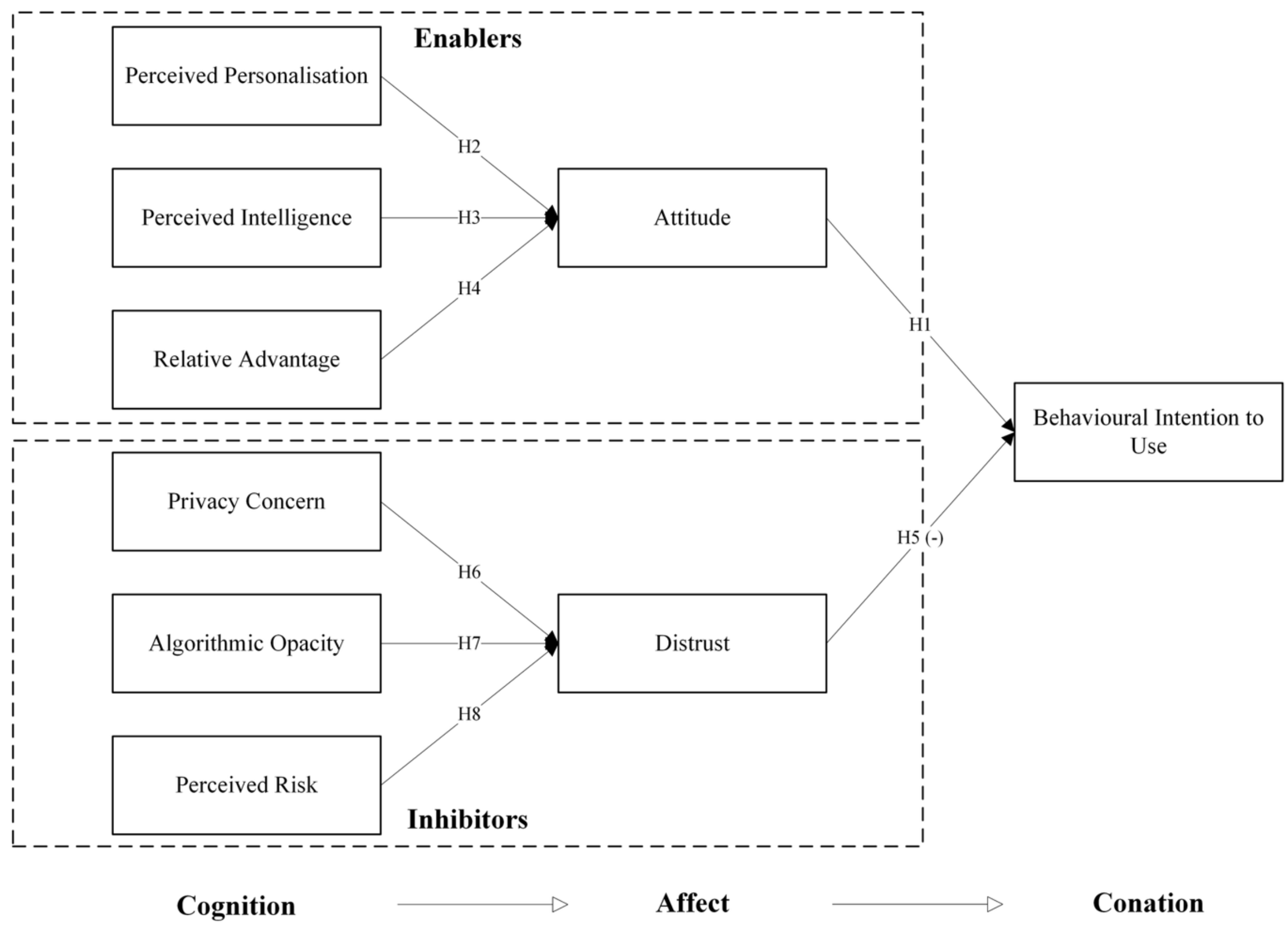

### 3. Hypothesis Development

#### 3.1 Impact of Attitude, Perceived Personalisation, Perceived Intelligence, and Relative Advantage

Attitude represents the affective evaluation formed after individuals cognitively assess a technology (Abbad, 2021). Prior research in information systems and technology acceptance consistently shows that attitude is a significant determinant of behavioural intention (Zheng et al., 2025a). Studies grounded in the Technology Acceptance Model and related frameworks demonstrate that when users develop favourable evaluations of a system, they are more likely to form intentions to use it (Granić, 2024). Empirical evidence from research on intelligent assistants and AI-based systems similarly indicates that positive attitudes significantly predict adoption intention, suggesting that users' emotional evaluation plays a central role in determining whether they intend to use agent-based technologies such as OpenClaw (Jung & Jo, 2025).

Perceived personalisation refers to the extent to which a system can adapt its content, services, or interactions to individual users' preferences and needs (Niu et al., 2026). Empirical studies on digital services, recommender systems, and AI-enabled platforms show that perceived personalisation enhances users' perceived relevance and satisfaction with a system. When users believe that a technology can deliver tailored responses or customised functionalities, they tend to evaluate the system more favourably (Law, 2024). Evidence from studies on personalised AI assistants and intelligent recommender systems suggests that personalisation improves users' affective responses toward technology, which subsequently contributes to more positive attitudes (Al Darayseh, 2023).

Perceived intelligence describes users' perception that a system can understand context, make appropriate decisions, and perform tasks competently (Tusseyeva et al., 2024). In the context of AI technologies, perceived intelligence has been shown to influence users' trust, perceived usefulness, and overall evaluation of the system (Chou et al., 2025). Empirical findings from research on conversational agents, smart assistants, and autonomous AI systems indicate that when users perceive a system as intelligent and capable, they tend to form stronger positive evaluations of the technology (Zhang et al., 2023). This perception enhances users' confidence in the system's performance and contributes to favourable attitudes (Zhou & Zhang, 2024).

Relative advantage refers to the degree to which a technology is perceived as offering superior benefits compared with existing alternatives (Choi, 2022). The concept originates from diffusion of innovation theory and has been widely validated as a predictor of technology adoption (Patnaik & Bakkar, 2024). Empirical studies across various information systems contexts demonstrate that users are more likely to adopt technologies that provide clear improvements in efficiency, productivity, or convenience (Setiawan & Alamsyah, 2022). When users perceive that OpenClaw offers functional advantages, such as automating complex tasks or integrating multiple tools into a single agentic system, they are more likely to develop a positive attitude toward the technology (Liang et al., 2025). Accordingly, the following hypotheses are proposed:

H1: Attitude positively influences behavioural intention to use OpenClaw.
H2: Perceived personalisation positively influences users' attitude toward OpenClaw.
H3: Perceived intelligence positively influences users' attitude toward OpenClaw.
H4: Relative advantage positively influences users' attitude toward OpenClaw.

**3.2 Impact of Distrust, Privacy Concern, Algorithmic Opacity, and Perceived Risk**

At the same time, negative cognitive evaluations may generate adverse affective responses that subsequently influence behavioural intention. In the context of AI-based systems, distrust represents a negative affective reaction reflecting scepticism, uncertainty, or lack of confidence in the technology (Peters & Visser, 2023). Prior research on technology acceptance and human–AI interaction indicates that distrust can significantly reduce users' willingness to rely on intelligent systems (Laux, 2024). Empirical studies on autonomous technologies and AI assistants show that when users perceive a system as unreliable or potentially harmful, distrust emerges and discourages adoption intentions (Afroogh et al., 2024).

Privacy concern refers to users' apprehension regarding the collection, use, and potential misuse of personal data by digital systems (Herriger et al., 2025). AI agents such as OpenClaw may require access to files, online services, or personal information in order to execute tasks, which may intensify users' concerns about data privacy (Carmody et al., 2021; Hu & Min, 2023). Empirical research in information systems consistently demonstrates that privacy concerns are negatively associated with users' trust and emotional evaluations of technologies (Zheng et al., 2025b). When individuals believe that a system may compromise their personal information, they tend to develop negative affective responses, including distrust (J. Li & Huang, 2020).

Algorithmic opacity describes the perceived lack of transparency in how an AI system processes information and makes decisions (Guo et al., 2025). Many advanced AI systems rely on complex algorithms that are difficult for users to interpret, which may create uncertainty regarding how outcomes are generated (Yang et al., 2024). Empirical evidence from studies on algorithmic decision-making suggests that low transparency often increases user scepticism and reduces confidence in automated systems (Vaassen, 2022). When users perceive AI processes as opaque or difficult to understand, they may question the system's fairness, reliability, and accountability, which can contribute to distrust (Eslami et al., 2019).

Perceived risk refers to users' expectation of potential negative outcomes associated with using a technology (W. Li, 2025). In the context of autonomous AI agents, risks may involve system errors, unintended actions, security vulnerabilities, or loss of control over automated processes (Wu et al., 2022). Prior research in technology adoption consistently identifies perceived risk as a significant barrier to acceptance (Goh et al., 2024). Empirical findings show that when users perceive higher levels of risk associated with a system, they tend to experience stronger negative affective reactions and are more likely to develop distrust toward the technology (Schwesig et al., 2023). Accordingly, the following hypotheses are proposed:

H5: Distrust negatively influences behavioural intention to use OpenClaw.
H6: Privacy concern positively influences users' distrust toward OpenClaw.
H7: Algorithmic opacity positively influences users' distrust toward OpenClaw.
H8: Perceived risk positively influences users' distrust toward OpenClaw.

## 4. Methods

### 4.1 Participants

Participants were recruited through the Credamo online survey platform, and the questionnaire link was also distributed on several online platforms and communities related to AI tools and automation technologies. To ensure that respondents could meaningfully evaluate the system, a screening question was used so that only individuals who had previously used OpenClaw were allowed to complete the questionnaire. After removing incomplete or invalid responses, a total of 436 valid questionnaires were retained for analysis. Among the respondents, 63.3% were male and 36.7% were female. In terms of age distribution, 53.2% were between 18 and 29 years old, 35.8% were between 30 and 44 years old, and 11.0% were aged 45 years or above. Regarding education level, 40.8% of participants held an undergraduate degree or lower, while 59.2% had a postgraduate degree. With respect to occupation, 42.2% of participants were students, 21.1% were academics or researchers, and 36.7% were industry professionals. Table 1 presents the detailed demographic characteristics of the respondents.

**Table 1. Demographic Characteristics of Respondents (*N* = 436)**

| Variable | Category | Frequency (*n*) | Percentage (%) |
|---|---|---|---|
| Age | 18–29 | 232 | 53.2 |
| | 30–44 | 156 | 35.8 |
| | 45 and above | 48 | 11.0 |
| Gender | Male | 276 | 63.3 |
| | Female | 160 | 36.7 |
| Education Level | Undergraduate or below | 178 | 40.8 |
| | Postgraduate | 258 | 59.2 |
| Occupation | Student | 184 | 42.2 |
| | Academic / Researcher | 92 | 21.1 |
| | Industry Professional | 160 | 36.7 |

### 4.2 Measurement

All measurement items were adapted from established studies in the information systems and technology adoption literature and modified to fit the context of OpenClaw (Y. Du, 2024). The measurement items were modified to fit the context of OpenClaw. Specifically, perceived personalisation, perceived intelligence, relative advantage, privacy concern, algorithmic opacity, and perceived risk were used to capture users' cognitive evaluations of the system, while attitude and distrust represented affective responses. Behavioural intention was measured to capture users' intention to continue using OpenClaw. All items were measured using a five-point Likert scale ranging from 1 (strongly disagree) to 5 (strongly agree). The detailed measurement items and their corresponding constructs are presented in Table 2.

**Table 2. Constructs and Measurement Items**

| Construct and Reference | Code | Measurement Item |
|---|---|---|
| Perceived Personalisation (PP) (Zhou & Zhang, 2024) | PP1 | OpenClaw provides responses tailored to my individual needs. |
| | PP2 | OpenClaw adjusts its services according to my preferences. |
| | PP3 | OpenClaw delivers personalised assistance when performing tasks. |
| Perceived Intelligence (PI) (Zhou & Wang, 2025) | PI1 | OpenClaw appears intelligent when executing tasks. |
| | PI2 | OpenClaw demonstrates an ability to understand my requests. |

| | PI3 | OpenClaw can make appropriate decisions when completing tasks. |
|---|---|---|
| Relative Advantage (RA) (Choi, 2022) | RA1 | Using OpenClaw improves the efficiency of completing tasks. |
| | RA2 | OpenClaw provides greater benefits than alternative tools I could use. |
| | RA3 | Using OpenClaw enhances my overall productivity. |
| Privacy Concern (PC) (Zhou & Zhang, 2025) | PC1 | I am concerned about the amount of personal information OpenClaw collects. |
| | PC2 | I am concerned about how OpenClaw uses my personal data. |
| | PC3 | I am concerned about the protection of my personal information when using OpenClaw. |
| Algorithmic Opacity (AO) (Guo et al., 2025) | AO1 | I find it difficult to understand how OpenClaw generates its outputs. |
| | AO2 | The decision-making process of OpenClaw is unclear to me. |
| | AO3 | It is hard to explain how OpenClaw arrives at its results. |
| Perceived Risk (PR) (Song & Zhou, 2026) | PR1 | Using OpenClaw involves potential risks. |
| | PR2 | I am concerned that OpenClaw may produce incorrect or harmful outcomes. |
| | PR3 | Relying on OpenClaw could lead to unexpected problems. |
| Attitude (ATT) (Y. Du et al., 2025) | ATT1 | Using OpenClaw is a good idea. |
| | ATT2 | I have a favourable opinion of OpenClaw. |
| | ATT3 | Overall, I like the idea of using OpenClaw. |
| Distrust (DT) (Shahzad et al., 2026) | DT1 | I feel sceptical about relying on OpenClaw. |
| | DT2 | I am uncertain about the reliability of OpenClaw. |
| | DT3 | I feel uneasy about depending on OpenClaw to perform tasks. |
| Behavioural Intention (BI) (C. Du et al., 2025) | BI1 | I intend to continue using OpenClaw in the future. |
| | BI2 | I plan to use OpenClaw whenever it is appropriate. |
| | BI3 | I will actively consider using OpenClaw for future tasks. |

**4.3 Data Analysis**

The data were analysed using structural equation modelling (SEM) (Whittaker & Schumacker, 2022) to examine the relationships among the study constructs and test the proposed hypotheses. Prior to hypothesis testing, descriptive statistics were computed to assess the distribution of the variables. Confirmatory factor analysis (CFA) was conducted to evaluate the measurement model, including model fit, reliability, and construct validity. Reliability was assessed using Cronbach's α and composite reliability (CR), while convergent validity was examined through factor loadings and average variance extracted (AVE). Discriminant validity was evaluated using the Fornell–Larcker criterion by comparing the square roots of AVE with inter-construct correlations. After establishing the adequacy of the measurement model, the structural model was estimated to test the hypothesised relationships among the constructs and evaluate the overall model fit.

**5. Results**

**5.1 Descriptive Statistics of the Constructs**

Table 3 presents the descriptive statistics of the study constructs. The mean values range from 2.94 to 3.92, indicating that respondents generally held moderately positive perceptions toward OpenClaw. Perceived intelligence ($M = 3.92$) and behavioural intention ($M = 3.90$) showed the highest mean values, followed by attitude ($M = 3.87$) and relative advantage ($M = 3.85$). In contrast, the inhibiting factors, privacy concern ($M = 3.21$), algorithmic opacity ($M = 3.34$), perceived risk ($M = 3.18$), and distrust ($M = 2.94$), exhibited relatively moderate levels. The standard deviations ranged from 0.68 to 0.82, suggesting acceptable variability among responses. Furthermore, the skewness and kurtosis values for all constructs fell within the acceptable range of ±1, indicating that the data distribution did not deviate substantially from normality.

**Table 3. Descriptive Statistics of the Constructs**

| Construct | *M* | SD | Skewness | Kurtosis |
|---|---|---|---|---|
| Perceived Personalisation (PP) | 3.78 | 0.71 | -0.41 | -0.22 |
| Perceived Intelligence (PI) | 3.92 | 0.69 | -0.53 | -0.18 |
| Relative Advantage (RA) | 3.85 | 0.73 | -0.47 | -0.26 |
| Privacy Concern (PC) | 3.21 | 0.82 | 0.28 | -0.39 |
| Algorithmic Opacity (AO) | 3.34 | 0.79 | 0.17 | -0.44 |
| Perceived Risk (PR) | 3.18 | 0.76 | 0.24 | -0.31 |
| Attitude (ATT) | 3.87 | 0.68 | -0.49 | -0.20 |
| Distrust (DT) | 2.94 | 0.81 | 0.36 | -0.27 |
| Behavioural Intention (BI) | 3.90 | 0.70 | -0.55 | -0.15 |

**5.2 Measurement Model**

The measurement model was evaluated using confirmatory factor analysis to assess model fit, reliability, and construct validity. As shown in Table 4, the measurement model demonstrated satisfactory overall fit to the data ($\chi^2/df = 2.14$, CFI = 0.94, TLI = 0.93, RMSEA = 0.051, SRMR = 0.046), meeting the recommended thresholds. Table 5 presents the reliability and convergent validity results. All factor loadings ranged from 0.79 to 0.92 and exceeded the recommended value of 0.70, indicating strong indicator reliability. Cronbach's α values ranged from 0.84 to 0.90 and composite reliability (CR) values ranged from 0.89 to 0.93, both above the recommended threshold of 0.70, confirming internal consistency reliability. The average variance extracted (AVE) values ranged from 0.72 to 0.81, exceeding the recommended threshold of 0.50, which indicates adequate convergent validity. Discriminant validity was assessed using the Fornell–Larcker criterion. As reported in Table 6, the square root of the AVE for each construct was greater than its correlations with other constructs, demonstrating satisfactory discriminant validity. Overall, the results confirm that the measurement model possesses adequate reliability and validity.

**Table 4. Model Fit Indices**

| Fit Index | Threshold | Measurement Model | Structural Model |
|---|---|---|---|
| $\chi^2/df$ | < 3.00 | 2.14 | 2.26 |
| CFI | > 0.90 | 0.94 | 0.93 |
| TLI | > 0.90 | 0.93 | 0.92 |
| RMSEA | < 0.08 | 0.051 | 0.054 |
| SRMR | < 0.08 | 0.046 | 0.049 |

**Table 5. Reliability and Convergent Validity**

| Construct | Item | Factor Loading | Cronbach's α | CR | AVE |
|---|---|---|---|---|---|
| Perceived Personalisation (PP) | PP1 | 0.79 | 0.86 | 0.90 | 0.74 |
| | PP2 | 0.88 | | | |
| | PP3 | 0.87 | | | |
| Perceived Intelligence (PI) | PI1 | 0.84 | 0.88 | 0.91 | 0.77 |
| | PI2 | 0.90 | | | |
| | PI3 | 0.88 | | | |
| Relative Advantage (RA) | RA1 | 0.81 | 0.87 | 0.90 | 0.75 |
| | RA2 | 0.89 | | | |
| | RA3 | 0.87 | | | |
| Privacy Concern (PC) | PC1 | 0.82 | 0.85 | 0.89 | 0.72 |
| | PC2 | 0.88 | | | |
| | PC3 | 0.83 | | | |
| Algorithmic Opacity (AO) | AO1 | 0.80 | 0.84 | 0.89 | 0.73 |
| | AO2 | 0.87 | | | |
| | AO3 | 0.86 | | | |
| Perceived Risk (PR) | PR1 | 0.83 | 0.86 | 0.90 | 0.75 |
| | PR2 | 0.89 | | | |
| | PR3 | 0.86 | | | |
| Attitude (ATT) | ATT1 | 0.88 | 0.89 | 0.92 | 0.79 |
| | ATT2 | 0.91 | | | |
| | ATT3 | 0.87 | | | |
| Distrust (DT) | DT1 | 0.82 | 0.85 | 0.89 | 0.73 |
| | DT2 | 0.88 | | | |
| | DT3 | 0.84 | | | |
| Behavioural Intention (BI) | BI1 | 0.89 | 0.90 | 0.93 | 0.81 |
| | BI2 | 0.92 | | | |
| | BI3 | 0.88 | | | |

**Table 6. Discriminant Validity (Fornell–Larcker Criterion)**

| Construct | PP | PI | RA | PC | AO | PR | ATT | DT | BI |
|---|---|---|---|---|---|---|---|---|---|
| PP | **0.86** | | | | | | | | |
| PI | 0.58 | **0.88** | | | | | | | |
| RA | 0.61 | 0.63 | **0.87** | | | | | | |
| PC | -0.21 | -0.24 | -0.27 | **0.85** | | | | | |
| AO | -0.18 | -0.20 | -0.23 | 0.46 | **0.86** | | | | |
| PR | -0.25 | -0.28 | -0.30 | 0.49 | 0.52 | **0.87** | | | |
| ATT | 0.66 | 0.69 | 0.71 | -0.29 | -0.25 | -0.31 | **0.89** | | |
| DT | -0.33 | -0.36 | -0.38 | 0.55 | 0.58 | 0.60 | -0.41 | **0.86** | |
| BI | 0.64 | 0.67 | 0.69 | -0.26 | -0.22 | -0.29 | 0.72 | -0.44 | **0.90** |

Note. Diagonal elements (bold) represent the square root of AVE. Off-diagonal elements represent the correlations among constructs.

### 5.3 Structural Model

The structural model was estimated to test the proposed hypotheses and examine the relationships among the constructs. As shown in Table 4, the structural model demonstrated acceptable overall model fit ($\chi^2$/df = 2.26, CFI = 0.93, TLI = 0.92, RMSEA = 0.054, SRMR = 0.049), indicating that the model adequately represents the observed data. The hypothesis testing results are presented in Table 7 and Figure 2. The results show that attitude had a significant positive effect on behavioural intention ($\beta$ = 0.49, $p < .001$), supporting H1, while distrust had a significant negative effect on behavioural intention ($\beta$ = −0.22, $p < .01$), supporting H5. Regarding the cognitive antecedents of attitude, perceived personalisation ($\beta$ = 0.21, $p < .001$), perceived intelligence ($\beta$ = 0.27, $p < .001$), and relative advantage ($\beta$ = 0.32, $p < .001$) all significantly and positively influenced attitude, supporting H2, H3, and H4. In addition, privacy concern ($\beta$ = 0.24, $p < .001$), algorithmic opacity ($\beta$ = 0.19, $p < .05$), and perceived risk ($\beta$ = 0.28, $p < .001$) had significant positive effects on distrust, supporting H6, H7, and H8. These results indicate that both enabling and inhibiting perceptions influence users' affective responses, which in turn shape their behavioural intention to use OpenClaw.

**Table 7. Structural Model Results**

| Hypothesis | Path | $\beta$ | SE | $z$ | Result |
|---|---|---|---|---|---|
| H1 | ATT → BI | 0.49 | 0.07 | 7.00*** | Supported |
| H2 | PP → ATT | 0.21 | 0.06 | 3.50*** | Supported |
| H3 | PI → ATT | 0.27 | 0.06 | 4.50*** | Supported |
| H4 | RA → ATT | 0.32 | 0.06 | 5.33*** | Supported |
| H5 | DT → BI | -0.22 | 0.07 | -3.14** | Supported |
| H6 | PC → DT | 0.24 | 0.07 | 3.43*** | Supported |
| H7 | AO → DT | 0.19 | 0.08 | 2.38* | Supported |
| H8 | PR → DT | 0.28 | 0.07 | 4.00*** | Supported |

Note. Statistical significance is denoted as *** $p < .001$, ** $p < .01$, * $p < .05$.

**Figure 2. Structural Model Results**

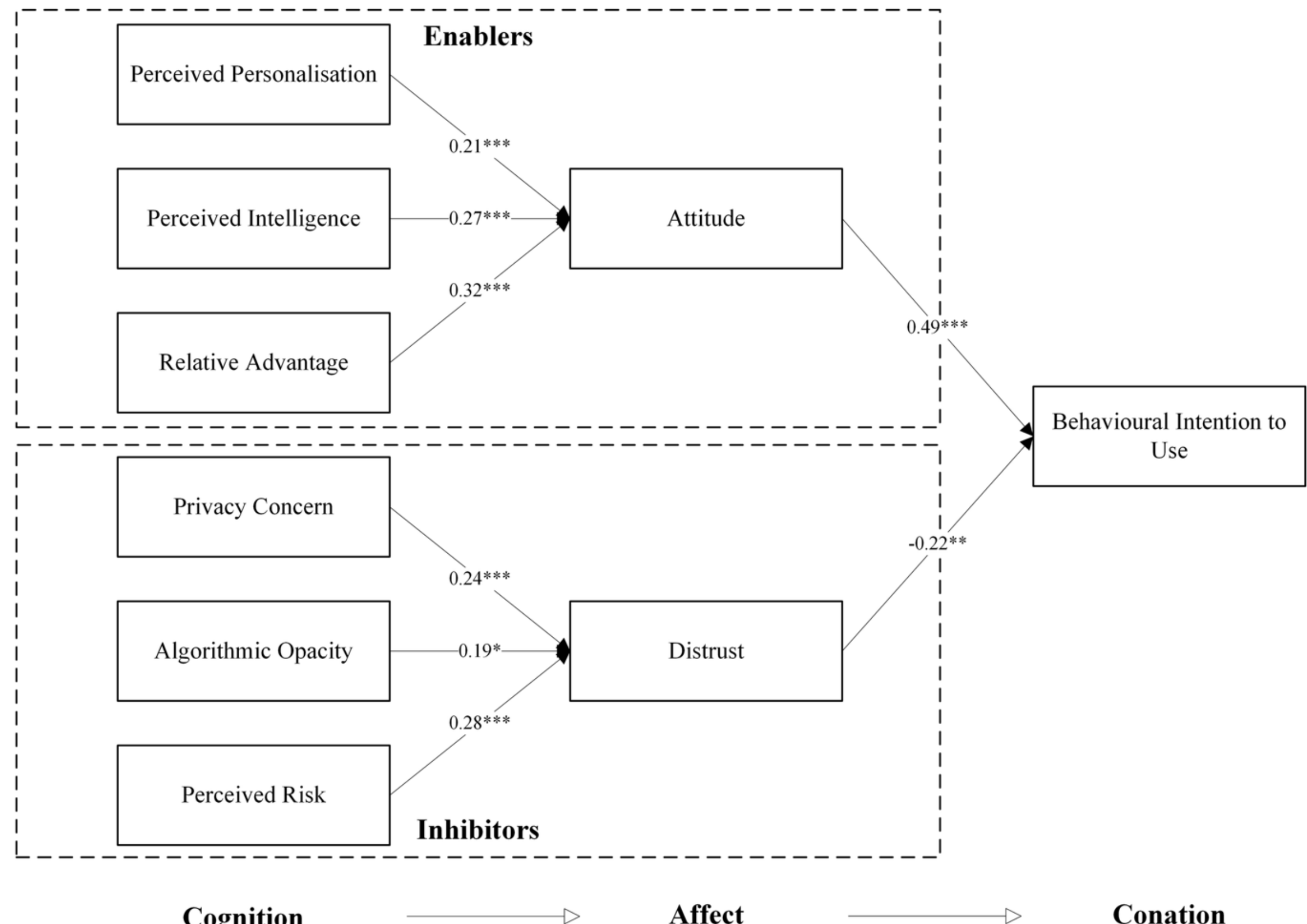

Note. Statistical significance is denoted as *** $p < .001$, ** $p < .01$, * $p < .05$.

## 6. Discussion

### 6.1 Impact of Attitude, Perceived Personalisation, Perceived Intelligence, and Relative Advantage

The findings indicate that attitude plays a critical role in shaping users' behavioural intention to use OpenClaw. In line with technology acceptance research, users who develop favourable evaluations of a technology are more likely to intend to adopt and continue using it. This result supports prior studies demonstrating that attitude is a central determinant of behavioural intention in the adoption of information systems and AI technologies (Abbad, 2021; Granić, 2024; Zheng et al., 2025a). From the perspective of the Cognition–Affect–Conation framework, the result confirms that affective responses act as an important mechanism through which users' perceptions of a technology translate into behavioural intentions (Zeng et al., 2023; Zhou & Zhang, 2024).

Perceived personalisation was also found to positively influence users' attitude toward OpenClaw. When users perceive that a system can tailor its responses and services according to their individual preferences and needs, they tend to evaluate the technology more favourably. This finding is consistent with existing literature suggesting that personalised features increase users' perceived relevance and satisfaction with AI-enabled systems (Law, 2024; Niu et al., 2026). Prior studies have similarly shown that personalisation strengthens users' affective responses toward intelligent technologies, thereby improving their overall attitude toward using such systems (Al Darayseh, 2023).

Perceived intelligence represents another important factor influencing users' attitude. When users believe that an AI system is capable of understanding requests, making appropriate decisions, and performing tasks effectively, they are more likely to form positive evaluations of the technology. This observation is consistent with research on human–AI interaction and conversational agents, which indicates that perceived intelligence enhances users' confidence in the system's capabilities and strengthens their positive perception of the technology (Chou et al., 2025; Tusseyeva et al., 2024; Zhang et al., 2023). In the context of agent-based systems such as OpenClaw, perceived intelligence is particularly important because users rely on the system to autonomously execute complex tasks.

Relative advantage also contributes significantly to the formation of positive attitudes toward OpenClaw. When users perceive that the system provides clear benefits compared with existing tools, such as improving efficiency, productivity, or task automation, they are more likely to evaluate the technology favourably. This finding aligns with diffusion of innovation theory, which identifies relative advantage as one of the most influential drivers of technology adoption (Choi, 2022; Patnaik & Bakkar, 2024). Previous studies on AI adoption have similarly demonstrated that perceived functional benefits enhance users' attitudes and encourage technology acceptance (Liang et al., 2025; Setiawan & Alamsyah, 2022). Collectively, these findings highlight the importance of users' cognitive evaluations of system capabilities in shaping affective responses and, ultimately, their intention to adopt autonomous AI agents such as OpenClaw.

### 6.2 Impact of Distrust, Privacy Concern, Algorithmic Opacity, and Perceived Risk

The results indicate that distrust negatively influences users' behavioural intention to use OpenClaw. When users feel uncertain about the reliability or safety of an AI system, they become less willing to rely on it for completing tasks. This finding is consistent with prior research suggesting that distrust acts as an important barrier to the adoption of AI technologies and automated systems (Peters & Visser, 2023; Laux, 2024). In the context of agentic AI systems, which can autonomously execute actions and interact with external tools, distrust may become particularly influential because users must rely on the system's decisions and operations. As a result, negative affective reactions can reduce users' willingness to adopt or continue using such technologies.

Privacy concern was found to positively influence distrust toward OpenClaw. When users worry about how their personal information may be collected, stored, or used by the system, they are more likely to develop sceptical or negative feelings toward the technology. This result aligns with existing research showing that privacy concerns significantly influence users' trust perceptions and attitudes toward digital systems (Carmody et al., 2021; Hu & Min, 2023). Studies in the AI adoption literature similarly indicate that concerns about data security and personal information protection can trigger negative emotional responses and increase users' distrust of AI-based systems (Herriger et al., 2025; Zheng et al., 2025b).

Algorithmic opacity also contributes to the development of distrust. When users perceive that the decision-making processes of an AI system are difficult to understand or lack transparency, they may question the fairness, reliability, or accountability of the system. This perception of opacity can create uncertainty about how outcomes are generated, which may lead to scepticism toward automated technologies. Prior research on algorithmic systems similarly suggests that low transparency often reduces users' confidence and increases negative perceptions of AI systems (Eslami et al., 2019; Vaassen, 2022). Consequently, improving transparency and explainability may be important for reducing distrust in agent-based AI technologies.

Perceived risk further strengthens users' distrust toward OpenClaw. When individuals believe that using an AI system may involve potential negative outcomes, such as system errors, unintended actions, or security vulnerabilities, they are more likely to feel uneasy about relying on it. This finding is consistent with prior technology adoption studies that identify perceived risk as a key factor influencing users' emotional reactions toward emerging technologies (Wu et al., 2022; Schwesig et al., 2023). In the context of autonomous AI agents, perceived risk may be particularly salient because the system performs tasks independently, which may increase users' concerns about loss of control or unintended consequences. Collectively, these results highlight how negative cognitive perceptions can generate distrust, which subsequently reduces users' intention to adopt AI agent technologies.

### 6.3 Theoretical and Practical Implications

From a theoretical perspective, this study contributes to the literature on AI adoption by applying the Cognition–Affect–Conation (CAC) framework to the context of autonomous AI agents such as OpenClaw. While previous research has primarily examined conversational AI or recommender systems, empirical studies on agent-based AI technologies remain limited. By integrating both enabling and inhibiting factors within a single framework, the study demonstrates how users' cognitive evaluations, such as perceived personalisation, perceived intelligence, relative advantage, privacy concern, algorithmic opacity, and perceived risk, shape affective responses, which subsequently influence behavioural intention. This finding extends prior CAC-based research in information systems by illustrating that both positive and negative perceptions jointly influence technology adoption decisions (Zeng et al., 2023; Zhou & Zhang, 2024; Zhou & Wang, 2025). Furthermore, the study contributes to the growing body of literature on human–AI interaction by highlighting the importance of both attitude and distrust as key affective mechanisms in users' decision-making regarding AI agents.

From a practical perspective, the findings provide important insights for developers and designers of agentic AI systems. Enhancing system capabilities that strengthen positive cognitive perceptions—such as improving personalisation, intelligent task execution, and functional advantages—can foster favourable attitudes and increase users' intention to adopt AI agents. At the same time, developers should address factors that contribute to distrust by improving transparency, strengthening data protection mechanisms, and reducing perceived risks associated with system autonomy. Designing AI systems that provide clearer explanations of decision-making processes and greater user control may help mitigate concerns related to algorithmic opacity and privacy. By balancing functional innovation with transparency and security considerations, developers can improve user confidence and facilitate the broader adoption of autonomous AI technologies.

### 6.4 Limitations and Future Direction

This study has several limitations that should be acknowledged. First, the data were collected from users who had prior experience with OpenClaw, which may limit the generalisability of the findings to broader populations or individuals who have not yet interacted with agent-based AI systems . Users with prior experience may already hold relatively informed perceptions about the system, which could influence their evaluations of its capabilities and risks. Additionally, the study employed a cross-sectional survey design, capturing participants' perceptions and intentions at a single point in time. Such a design may not fully reflect how users' attitudes and perceptions evolve as they gain more experience with autonomous AI technologies.

Future research could extend this study in several directions. Longitudinal studies would be valuable for examining how users' perceptions, attitudes, and behavioural intentions change over time as individuals interact more extensively with AI agents. In addition, future research could incorporate additional constructs, such as trust, perceived usefulness, social influence, or algorithmic literacy, to provide a more comprehensive understanding of AI adoption behaviour. Researchers may also consider employing qualitative or mixed-method approaches to explore users' experiences with agent-based AI systems in greater depth, which could generate richer insights into how individuals interpret, evaluate, and interact with emerging autonomous AI technologies.

## 7. Conclusion

This study examined users' behavioural intention to use OpenClaw through the Cognition–Affect–Conation framework by analysing how cognitive perceptions shape affective responses and ultimately influence adoption intentions. The findings indicate that positive perceptions of the system, specifically perceived personalisation, perceived intelligence, and relative advantage, contribute to favourable attitudes toward OpenClaw, which subsequently strengthen users' intention to use the technology. At the same time, negative perceptions such as privacy concern, algorithmic opacity, and perceived risk increase users' distrust, which in turn reduces behavioural intention to adopt the system. These results highlight the importance of both enabling and inhibiting perceptions in shaping users' responses to autonomous AI agents. Overall, the study provides empirical insights into the psychological mechanisms underlying the adoption of agent-based AI technologies and offers implications for designing AI systems that balance functional capabilities with transparency, privacy protection, and user trust.

**Reference**

Abbad, M. M. M. (2021). Using the UTAUT model to understand students' usage of e-learning systems in developing countries. *Education and Information Technologies*, *26*(6), 7205–7224. https://doi.org/10.1007/s10639-021-10573-5

Afroogh, S., Akbari, A., Malone, E., Kargar, M., & Alambeigi, H. (2024). Trust in AI: Progress, challenges, and future directions. *Humanities and Social Sciences Communications*, *11*(1), 1568. https://doi.org/10.1057/s41599-024-04044-8

Al Darayseh, A. (2023). Acceptance of artificial intelligence in teaching science: Science teachers' perspective. *Computers and Education: Artificial Intelligence*, *4*, 100132. https://doi.org/10.1016/j.caeai.2023.100132

Basu, M. (2026). OpenClaw AI chatbots are running amok—These scientists are listening in. *Nature*, *650*(8102), 533–534. https://doi.org/10.1038/d41586-026-00370-w

Borjigin, A., Stadnyk, I., Bilski, B., Hovorov, S., & Pidturkina, S. (2026). *Execution is the new attack surface: Survivability-aware agentic crypto trading with OpenClaw-style local executors* (arXiv:2603.10092). arXiv. https://doi.org/10.48550/arXiv.2603.10092

Carmody, J., Shringarpure, S., & Van De Venter, G. (2021). AI and privacy concerns: A smart meter case study. *Journal of Information, Communication and Ethics in Society*, *19*(4), 492–505. https://doi.org/10.1108/JICES-04-2021-0042

Chen, E., Guan, C., Elshafiey, A., Zhao, Z., Zekeri, J., Shaibu, A. E., & Prince, E. O. (2026). *When OpenClaw AI agents teach each other: Peer learning patterns in the moltbook community* (Version 1). arXiv. https://doi.org/10.48550/ARXIV.2602.14477

Chen, T., Liu, D., Hu, X., Yu, J., & Wang, W. (2026). *A trajectory-based safety audit of clawdbot (OpenClaw)* (Version 1). arXiv. https://doi.org/10.48550/ARXIV.2602.14364

Choi, J. (2022). Enablers and inhibitors of smart city service adoption: A dual-factor approach based on the technology acceptance model. *Telematics and Informatics*, *75*, 101911. https://doi.org/10.1016/j.tele.2022.101911

Chou, C.-M., Shen, T.-C., Shen, T.-C., & Shen, C.-H. (2025). Teachers' adoption of AI-supported teaching behavior and its influencing factors: Using structural equation modeling. *Journal of Computers in Education*, *12*(3), 853–896. https://doi.org/10.1007/s40692-024-00332-z

Dong, B., Feng, H., & Wang, Q. (2026). *Clawdrain: Exploiting tool-calling chains for stealthy token exhaustion in OpenClaw agents* (Version 1). arXiv. https://doi.org/10.48550/ARXIV.2603.00902

Du, C., Tang, M., Wang, C., Zou, B., Xia, Y., & Du, Y. (2025). Who is most likely to accept AI chatbots? A sequential explanatory mixed-methods study of personality and ChatGPT acceptance for language learning. *Innovation in Language Learning and Teaching*, 1–22. https://doi.org/10.1080/17501229.2025.2555515

Du, Y. (2024). A streamlined approach to scale adaptation: Enhancing validity and feasibility in educational measurement. *Journal of Language Teaching*, *4*(3), 18–22. https://doi.org/10.54475/jlt.2024.017

Du, Y., Wang, C., Zou, B., & Xia, Y. (2025). Personalizing AI tools for second language speaking: The role of gender and autistic traits. *Frontiers in Psychiatry*, *15*, 1464575. https://doi.org/10.3389/fpsyt.2024.1464575

Eslami, M., Vaccaro, K., Lee, M. K., Elazari Bar On, A., Gilbert, E., & Karahalios, K. (2019). User attitudes towards algorithmic opacity and transparency in online reviewing platforms. *Proceedings of the 2019 CHI Conference on Human Factors in Computing Systems*, 1–14. https://doi.org/10.1145/3290605.3300724

Goh, W. W., Chia, K. Y., Cheung, M. F., Kee, K. M., Lwin, M. O., Schulz, P. J., Chen, M., Wu, K., Ng, S. S., Lui, R., Ang, T. L., Yeoh, K. G., Chiu, H., Wu, D., & Sung, J. J. (2024). Risk perception, acceptance, and trust of using AI in gastroenterology practice in the asia-pacific region: Web-based survey study. *JMIR AI*, *3*, e50525. https://doi.org/10.2196/50525

Granić, A. (2024). Technology adoption at individual level: Toward an integrated overview. *Universal Access in the Information Society*, *23*(2), 843–858. https://doi.org/10.1007/s10209-023-00974-3

Guo, C., Liu, H., Song, F., & Guo, J. (2025). The double-edged sword effects of algorithmic opacity: The self-determination theory perspective. *Acta Psychologica*, *260*, 105600. https://doi.org/10.1016/j.actpsy.2025.105600

Herriger, C., Merlo, O., Eisingerich, A. B., & Arigayota, A. R. (2025). Context-contingent privacy concerns and exploration of the privacy paradox in the age of AI, augmented reality, big data, and the internet of things: Systematic review. *Journal of Medical Internet Research*, *27*, e71951. https://doi.org/10.2196/71951

Hu, Y., & Min, H. (Kelly). (2023). The dark side of artificial intelligence in service: The "watching-eye" effect and privacy concerns. *International Journal of Hospitality Management*, *110*, 103437. https://doi.org/10.1016/j.ijhm.2023.103437

Jung, Y. M., & Jo, H. (2025). Understanding continuance intention of generative AI in education: An ECM-based study for sustainable learning engagement. *Sustainability*, *17*(13), 6082. https://doi.org/10.3390/su17136082

Laux, J. (2024). Institutionalised distrust and human oversight of artificial intelligence: Towards a democratic design of AI governance under the European Union AI Act. *AI & SOCIETY*, *39*(6), 2853–2866. https://doi.org/10.1007/s00146-023-01777-z

Law, L. (2024). Application of generative artificial intelligence (GenAI) in language teaching and learning: A scoping literature review. *Computers and Education Open*, *6*, 100174. https://doi.org/10.1016/j.caeo.2024.100174

Li, J., & Huang, J.-S. (2020). Dimensions of artificial intelligence anxiety based on the integrated fear acquisition theory. *Technology in Society*, *63*, 101410. https://doi.org/10.1016/j.techsoc.2020.101410

Li, W. (2025). A study on factors influencing designers' behavioral intention in using AI-generated content for assisted design: Perceived anxiety, perceived risk, and UTAUT. *International Journal of Human–Computer Interaction*, *41*(2), 1064–1077. https://doi.org/10.1080/10447318.2024.2310354

Liang, J., Zhu, Y., Wu, J., & Chen, C. (2025). "When I have the advantage, I prefer AI!" the influence of an applicant's relative advantage on the preference for artificial intelligence decision-making. *Journal of Business and Psychology*, *40*(5), 1209–1229. https://doi.org/10.1007/s10869-025-10012-z

Manik, M. M. H., & Wang, G. (2026). *OpenClaw agents on moltbook: Risky instruction sharing and norm enforcement in an agent-only social network* (Version 1). arXiv. https://doi.org/10.48550/ARXIV.2602.02625

Niu, J. W., Wang, K., Wang, L., Ruan, W.-Q., & Xiao, H. (2026). Resistance to AI-designed customized travel: The role of perceived personalization. *Tourism Review*, *81*(2), 651–668. https://doi.org/10.1108/TR-09-2024-0824

Patnaik, P., & Bakkar, M. (2024). Exploring determinants influencing artificial intelligence adoption, reference to diffusion of innovation theory. *Technology in Society*, *79*, 102750. https://doi.org/10.1016/j.techsoc.2024.102750

Peters, T. M., & Visser, R. W. (2023). The importance of distrust in AI. In L. Longo (Ed.), *Explainable Artificial Intelligence* (Vol. 1903, pp. 301–317). Springer Nature Switzerland. https://doi.org/10.1007/978-3-031-44070-0_15

Qaisar, S., Nawaz Kiani, A., & Jalil, A. (2024). Exploring discontinuous intentions of social media users: A cognition-affect-conation perspective. *Frontiers in Psychology*, *15*, 1305421. https://doi.org/10.3389/fpsyg.2024.1305421

Schwesig, R., Brich, I., Buder, J., Huff, M., & Said, N. (2023). Using artificial intelligence (AI)? Risk and opportunity perception of AI predict people's willingness to use AI. *Journal of Risk Research*, *26*(10), 1053–1084. https://doi.org/10.1080/13669877.2023.2249927

Setiawan, S., & Alamsyah, D. P. (2022). Mediation model of relative advantage in mobile payment. *2022 International Conference on Decision Aid Sciences and Applications (DASA)*, 71–75. https://doi.org/10.1109/DASA54658.2022.9765112

Shahzad, K., Khan, A. N., Ahmad, B., Hayat, K., & Chang, S. (2026). Balancing trust and distrust in generative AI chatbot adoption: A case study from China. *The Service Industries Journal*, *46*(3–4), 308–331. https://doi.org/10.1080/02642069.2025.2487819

Shan, Z., Xin, J., Zhang, Y., & Xu, M. (2026). *Don't let the claw grip your hand: A security analysis and defense framework for OpenClaw* (arXiv:2603.10387). arXiv. https://doi.org/10.48550/arXiv.2603.10387

Song, C., & Zhou, S. (2026). Push or Pull? Understanding switching intentions from human services to GAI agents through the PPM framework. *Computers in Human Behavior*, *176*, 108874. https://doi.org/10.1016/j.chb.2025.108874

Tusseyeva, I., Sandygulova, A., & Rubagotti, M. (2024). Perceived intelligence in human-robot interaction: A review. *IEEE Access*, *12*, 151348–151359. https://doi.org/10.1109/ACCESS.2024.3478751

Vaassen, B. (2022). AI, opacity, and personal autonomy. *Philosophy & Technology*, *35*(4), 88. https://doi.org/10.1007/s13347-022-00577-5

Whittaker, T. A., & Schumacker, R. E. (2022). *A beginner's guide to structural equation modeling* (Fifth edition). Routledge.

Wu, W., Zhang, B., Li, S., & Liu, H. (2022). Exploring factors of the willingness to accept AI-assisted learning environments: An empirical investigation based on the UTAUT model and perceived risk theory. *Frontiers in Psychology*, *13*, 870777. https://doi.org/10.3389/fpsyg.2022.870777

Yang, H., Li, D., & Hu, P. (2024). Decoding algorithm fatigue: The role of algorithmic literacy, information cocoons, and algorithmic opacity. *Technology in Society*, *79*, 102749. https://doi.org/10.1016/j.techsoc.2024.102749

Zeng, S., Lin, X., & Zhou, L. (2023). Factors affecting consumer attitudes towards using digital media platforms on health knowledge communication: Findings of cognition–affect–conation pattern. *Frontiers in Psychology*, *14*, 1008427. https://doi.org/10.3389/fpsyg.2023.1008427

Zhang, C., Schießl, J., Plößl, L., Hofmann, F., & Gläser-Zikuda, M. (2023). Acceptance of artificial intelligence among pre-service teachers: A multigroup analysis. *International Journal of Educational Technology in Higher Education*, *20*(1), 49. https://doi.org/10.1186/s41239-023-00420-7

Zheng, W., Ma, Z., Sun, J., Wu, Q., & Hu, Y. (2025a). Exploring factors influencing continuance intention of pre-service teachers in using generative artificial intelligence. *International Journal of Human–Computer Interaction*, *41*(16), 10325–10338. https://doi.org/10.1080/10447318.2024.2433300

Zheng, W., Ma, Z., Sun, J., Wu, Q., & Hu, Y. (2025b). Exploring factors influencing continuance intention of pre-service teachers in using generative artificial intelligence. *International Journal of Human–Computer Interaction*, *41*(16), 10325–10338. https://doi.org/10.1080/10447318.2024.2433300

Zhou, T., & Wang, M. (2025). Examining generative AI user discontinuance from a dual perspective of enablers and inhibitors. *International Journal of Human–Computer Interaction*, *41*(20), 13140–13150. https://doi.org/10.1080/10447318.2025.2470280

Zhou, T., & Zhang, C. (2024). Examining generative AI user addiction from a C-A-C perspective. *Technology in Society*, *78*, 102653. https://doi.org/10.1016/j.techsoc.2024.102653

Zhou, T., & Zhang, C. (2025). Examining generative AI user intermittent discontinuance from a C-A-C perspective. *International Journal of Human–Computer Interaction*, *41*(10), 6377–6387. https://doi.org/10.1080/10447318.2024.2376370